# Unsupervised Domain Adaptation Learning Algorithm for RGB-D Staircase Recognition

JING Wang[1], KUANGEN Zhang[1, 2]

*(1. Department of Mechanical Engineering, The University of British Columbia, Vancouver V6T1Z4;
2. Department of Mechanical and Energy Engineering, Southern University of Science and Technology, Shenzhen 518055)*

**Abstract:** Detection and recognition of staircase as upstairs, downstairs and negative (e.g., ladder, level ground) are the fundamentals of assisting the visually impaired to travel independently in unfamiliar environments. Previous researches have focused on using massive amounts of RGB-D scene data to train traditional machine learning (ML) based models to detect and recognize the staircase. However, the performance of traditional ML techniques is limited by the amount of labeled RGB-D staircase data. In this paper, we apply an unsupervised domain adaptation approach in deep architectures to transfer knowledge learned from the labeled RGB-D escalator staircase dataset to the unlabeled RGB-D stationary dataset. By utilizing the domain adaptation method, our feedforward convolutional neural networks (CNN) based feature extractor with 5 convolution layers can achieve 100% classification accuracy on testing the labeled escalator staircase data distributions and 80.6% classification accuracy on testing the unlabeled stationary data distributions. We demonstrate the success of the approach for classifying staircase on two domains with a limited amount of data. To further demonstrate the effectiveness of the approach, we also validate the same CNN model without domain adaptation and compare its results with those of our proposed architecture.

## 1 Introduction

Approximately 285 million people around the world are visually impaired, of whom 39 million were blind, based on the 2010 World Health Organization survey [1]. Although some mobility assistant systems, which are based on converting sonar information into audible signals, have been developed to facilitate navigation, obstacle detection, and wayfinding tasks [2], the visually impaired still face challenges to actively interact with the dynamic surrounding environment, such as staircases. Also, Staircases widely exist in both the indoor and outdoor environment. Thus, the potential risk of falling from the stairs, especially the downstairs case, is fatal to the visually impaired.

Monocular cameras, stereo cameras, and laser scanning devices (e.g., LiDAR) have been used for detecting staircases. However, the existing approaches can only provide user-selected information instead of taking into account the obstacles presented in motion autonomously [8]. Recently, the RGB-D cameras, which can capture both visual features and depth information of the environment simultaneously, are widely used to recognize indoor and outdoor environments [3-4]. Many researchers have focused on combining visual features with depth information for robust image representation [5-9]. He [9] directly combined RGB channels and the depth channel together to generate four-channel RGBD images. Through feeding their RGBD images in a feedforward CNN, they obtained a higher validation accuracy compared with that of feeding original RGB images in the same CNN architecture. The research implies that depth channel has a better feature representation than R, G, B channels. However, their training and testing samples come from the same data distribution. If their model is used for testing another dataset, the performance will degrade significantly [10].

Munoz et al. [8] and Wang et al. [7] constructed one-dimensional depth vectors (parallel lines), which contains the distance from camera and orientation of each stair, to train a support vector machine (SVM) multi-classifier to classify upstairs, downstairs and level ground cases. They both used edge detection to gain edge maps from RGB images and extracted one-dimensional depth vectors with the distance and orientation information from edge maps by Hough transform. Nevertheless, the performance of edge detection algorithms and Hough transform highly depends on the level of thresholds chosen. This means they carried out numerous preprocessing tasks to make their datasets adapt to their model. Thus, their model will poorly perform when it is tested on real-case scenarios, whose data distribution is different from that of their own dataset.

Domain adaptation (DA) is a scenario when we aim at learning a discriminative classifier from a source data distribution and applying this classifier to a different but related target data distribution. DA approaches are able to learn a mapping between the source and the target domains when the target domain data is either fully unlabeled (unsupervised domain adaptation) or partially labeled (semi-supervised domain adaptation). In this work, we propose a new framework for recognizing staircase images in the presence of a shift (as shown in Fig.1) between training (source domain) and test (target domain) distribution. We formulate our task as an unsupervised domain adaptation (UDA) problem, in which we have the labeled escalator staircase visual features and depth features in the source domain while we have the **unlabeled** stationary staircase visual features and depth features in the target domain.

In order to apply our feed-forward network to the stationary staircase (target) domain without being hindered by the shift between the two domains, we need to embed domain adaptation into the process of learning representation so that the classification decisions are made based on features that are both discriminative and invariant to the change of domains. Thus, we decide to use domain adversarial neural networks (DANN) [11], which embeds feature extracting learning, label predicting learning and domain classifying learning into a composed deep feedforward network, to classify the staircase images across two domains. DANN achieves unsupervised domain adaptation by adding a domain classifier connected to the feature extractor via a gradient reversal layer. The gradient reversal layer makes extracted features become domain invariant by multiplying the gradient by -1 during the backpropagation. However, DANN experiments on MNIST and MNIST-M datasets, which have much fewer features and less noise compared with our staircase datasets. Their feedforward feature extractor is not deep enough to robustly extract necessary features from staircase images. Thus, we replace the original feature extractor with a deep CNN based architecture.

We summarize the main contributions of this paper as follows: 1) we verify the possibility of using domain adaptation methods to train a single model that can be applied to recognize staircase images with different data distributions; 2) We make a 4-channel RGB-D image dataset based on RGB-D Staircase Detection Dataset [8].

The remainder of the paper is organized as follows: Section 2 discusses the proposed framework and experimental methods of the research. Section 3 presents the experimental results. Section 4 discusses the results of the research. Section 5 concludes the paper and presents future work.

## 2 Unsupervised Domain Adaptation Learning

For classification tasks, we have the input space X and the set of L possible labels Y. In unsupervised domain adaptation, we sample n labeled i.i.d examples from the escalator staircase distribution $P(X^s, Y^s)$ to form the source domain $D_s = \{(X_i^s, Y_i^s)\}_{i=1}^n$. Similarly, we sample m **unlabeled** i.i.d examples from the stationary staircase data distribution $Q(X^t, Y^t)$, which is different but similar to the escalator staircase data distribution, to form the target domain $D_t = \{(X_j^t)\}_{j=1}^m$.

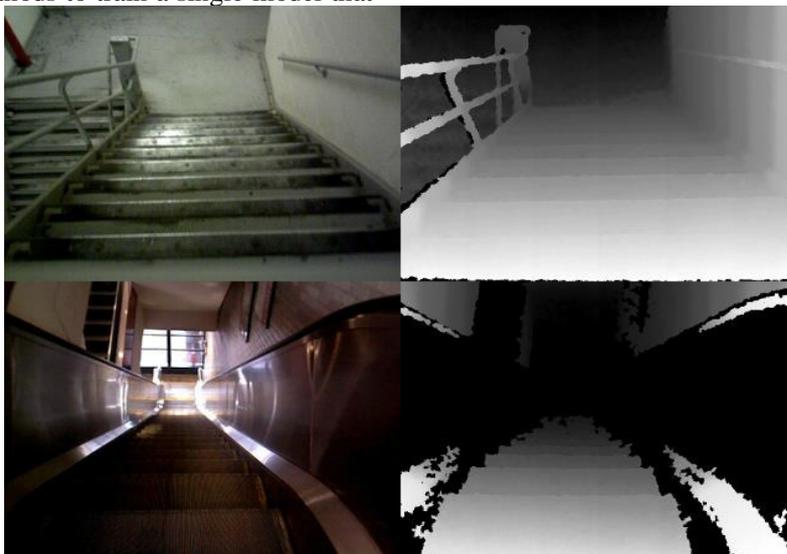

Fig.1 Examples of the RGB image and the depth image of the downstairs stationary staircase dataset are shown on the top-left and top-right corners individually. Examples of the RGB image and the depth image of the downstairs escalator staircase dataset are shown on the down-left and down-right corners individually. An obvious shift between two data distributions can be observed.

The work of Ben-David et al. showed that the empirical target risk $R_T$ is upper bounded by the empirical source risk $R_S$ plus the divergence of the hypothesis space $d_H(S,T)$ and a constant complexity term that depends on the VC-dimension d and the size of samples S and T [12].
**Theorem 1** [12] *Let H be a hypothesis space of VC-dimension d. With probability at least 1-δ (over the choice of samples $S \sim D_S$ and $T \sim D_T$), for every $h \in H$:*

$$R_T \leq R_S + d_H(S,T) + \frac{4}{m}\sqrt{\left(dlog\frac{2em}{d} + log\frac{4}{\delta}\right)} + 4\sqrt{\frac{dlog(2m)+log\frac{4}{\delta}}{m}}$$

The theorem indicates that the learning algorithm should minimize a trade-off between the empirical source risk and the empirical divergence $d_H(S,T)$ to minimize the empirical target risk. In this research, we aim to learn a feature extractor $f = G_f(x, \theta_f)$ that extracts domain-invariant features, a label predictor $y = G_y(f, \theta_y)$ that computers label predictions, and a domain classifier $G_d(f, \theta_d)$ that computers domain predictions and reduce the shift in two joint distributions. So, the empirical target risk $R_T = \frac{1}{m}\sum_{i=1}^m Pr_{(x,y) \sim Q}\left[G_y\left(G_f(x_i)\right) \neq y_i\right]$ can be minimized by jointly minimizing the source risk and the distribution difference.

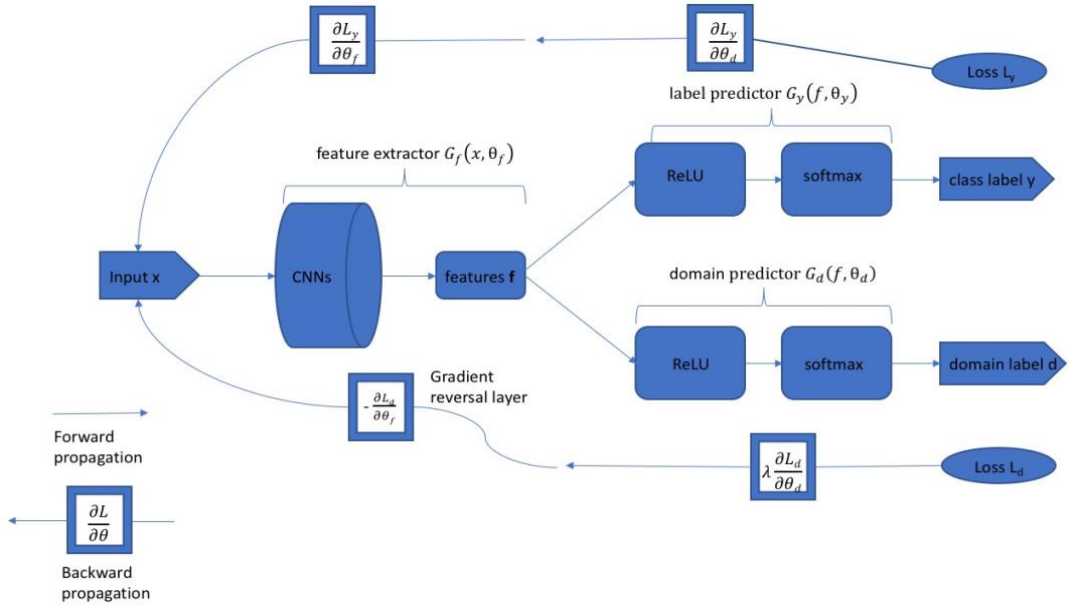

Fig.2    Domain adversarial neural network includes a standard feed-forward architecture formed by a deep feature extractor and a deep label predictor. The shift in two joint distributions is reduced by adding a domain predictor connected to the feature extractor via a gradient reversal layer. The gradient reversal layer multiplies the gradient by negative 1 during the backpropagation training, which makes the feature distributions over the two domains become similar.

### 2.1 Domain Adversarial Neural Networks

Domain adversarial neural networks explicitly implement the Theorem 1 into a neural network classifier that can extract transferable features to reduce the distribution shift between two domains. In order to extract as many transferable features as possible, we design a deep convolutional neural network $G_f(.,\theta_f)$, with parameters $\theta_f$, for learning discriminative features of images [14]. In this research, we use cross-entropy loss with softmax function to define the distance between predictions and the original input distributions. To ensure that the extracted features are domain-invariant, the parameter $\theta_f$ is jointly learned by maximizing the loss of domain predictor $G_d(f, \theta_d)$ and minimizing the loss of label predictor $G_y(f, \theta_y)$. We denote the label prediction loss and the domain prediction loss respectively by:

$$L_y(\theta_f, \theta_y) = L_y(G_y(G_f(x_i,\theta_f),\theta_y), y_i) \quad (1)$$
$$L_d(\theta_f, \theta_d) = L_d(G_d(G_f(x_i,\theta_f),\theta_d), d_i) \quad (2)$$

Thus, the domain adaptation problem becomes optimizing the objective function

$$R(\theta_f, \theta_y, \theta_d) = \frac{1}{n_s}\sum_{x_i \in D_S} L_y(\theta_f, \theta_y) - \frac{\lambda}{n_t+n_s}\sum_{x_i \in (D_S \cup D_T)} L_d(\theta_f, \theta_d) \quad (3)$$

Where $n_s$ and $n_t$ are the number of examples in the source domain and target domain respectively; and the loss of the domain classifier is weight by $\lambda$. By optimizing the objective function (3), the parameters $\theta_f, \theta_y, \theta_d$ will deliver a saddle point $(\hat{\theta}_f, \hat{\theta}_y, \hat{\theta}_d)$

$$(\hat{\theta}_f, \hat{\theta}_y) = \underset{\theta_f,\theta_y}{\arg\min}\, R(\theta_f, \theta_y, \hat{\theta}_d), \quad (4)$$
$$(\hat{\theta}_d) = \underset{\theta_d}{\arg\max}\, R(\hat{\theta}_f, \hat{\theta}_y, \theta_d). \quad (5)$$

by the following gradient updates:

$$\theta_f - \mu(\frac{\partial L_y}{\partial \theta_f} - \frac{\partial L_d}{\partial \theta_d}) \to \theta_f \quad (6)$$
$$\theta_y - \mu(\frac{\partial L_y}{\partial \theta_y}) \to \theta_y \quad (7)$$
$$\theta_d - \mu\lambda(\frac{\partial L_d}{\partial \theta_d}) \to \theta_d \quad (8)$$

Where $\mu$ is the learning rate.

### 2.2 CNN Architecture for Feature Extractor

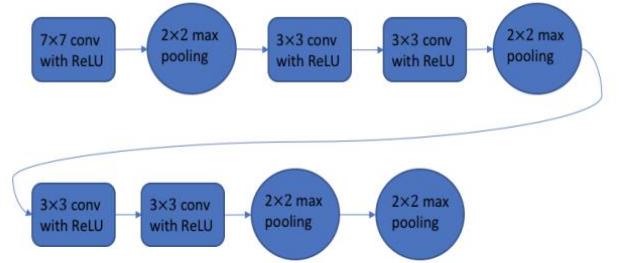

Fig.3    The proposed deep convolutional neural network architecture for feature extractor. The input of the feedforward architecture consists of 80×80× 4 RGB-D images. The output is a 1200-dimensional ($5 \times 5 \times 48$) descriptor which contains discriminative features.

To perform domain adversarial training, we construct a deep convolutional neural network for feature extractor $G_f(.,\theta_f)$. Ganin et al. experimented their domain adversarial neural network on MNIST and MNIST-M datasets. The two data distributions have much fewer features and less noise compared with our staircase datasets. Thus, our feature extractor should be deep enough, which can extract and learn discriminative features of staircase datasets.

The feature extractor $G_f(.,\theta_f)$ incorporates five convolution layers and one fully-connected layer, which gives 1200-dimensional descriptor as an output.

The first convolution layer applies 32 7×7 filters (extracting 7×7-pixel sub-regions), with ReLU activation function and 2×2 max pooling. The second convolution layer applies 32 3×3 filters with ReLU activation function. The third convolution layer applies 48 3×3 filters, with ReLU activation function and 2×2 max pooling. The fourth convolution layer applies 48 3×3 filters with ReLU activation function. The fifth convolution layer applies 48 3×3 filters, with ReLU activation function, followed by two 2×2 max pooling layers. During each training batch, the pairwise cosine similarities between 1200-dimensional features will be calculated and back-propagated as the loss for all pairs within the batch.

A label predictor $G_y(f, \theta_y)$ and a domain predictor $G_d(f, \theta_d)$ are attached to the 1200-dimensional bottleneck of the fully-connected layer of feature extractor in parallel (as shown in Fig.2). The label predictor consists of three fully connected layers (1200 → 100 → 100 → 3). Similarly, the domain predictor has two fully-connected layers (1200 → 100 → 2). $\theta_f$ is learned by jointly training the label predictor and the domain predictor. Thus, domain-invariant features can be effectively extracted by our feature extractor after training convergence.

### 2.3 RBG-D Data Generation

As we discussed above, depth information has a better feature representation than R, G, B channels. So, following He et al. we extend the original 3-channel RGB images to 4 channels, and simply copy the corresponding depth information to the fourth channel.

# 3 Results

We selected 1105 escalator RGB-D image pairs and 2157 stationary RGB-D image pairs from RGB-D Staircase Detection Dataset [8], which consists of three different categories (downstairs, upstairs, and negative cases). In order to combine RGB images with the corresponding depth information, we resized all RGB images to 72×72×4, and copied the corresponding depth information to their fourth channel. The escalator staircase dataset (source domain) was split into a training set (80%) and a testing set (20%). The stationary staircase dataset (target domain) was split into a training set (40%) and a testing set (60%). Each data sample is a 72×72×4 numpy array. We adjusted the learning rate during the stochastic gradient descent by $\mu = \frac{\mu_0}{(1+\alpha \cdot p)^\beta}$. The learning rate could alternatively be adjusted by cosine decay $\mu = \frac{1}{2}\left(1 + \cos\frac{t\pi}{T}\right)\mu_0$ at batch t, where T is the total number of batches [17]. During the training process, we set the initial learning rate $\mu_0$, momentum p, and batch size as 0.0005, 0.45, and 256, respectively. In order to promote convergence and low error on the source domain training, we set $\alpha = 10$ and $\beta = 0.75$. The domain adaptation parameter $\lambda$ is initiated at 0. During the training process, $\lambda$ is updated through $\lambda = \frac{2}{1+\exp(-10p)} - 1$.

### 3.1 Different Staircases

Three types of the staircase are shown in Fig.4. We can easily distinguish the difference between different types of the staircase. By comparing the staircase RGB image with its corresponding depth information, we argue that depth information has better feature representation and less noise than RGB information. Thus, more discriminative features can be extracted from the depth channel rather than from R, G, B channels.

### 3.2 Results of Testing the Proposed CNN Architecture without Domain Adaptation Regularizer

When relatively large amounts of training data are available, CNN can learn more discriminative features than any other existed methods do [15]. In order to show the advantage of using CNN to extract discriminative features, we trained and tested our proposed CNN architecture on the escalator staircase dataset (source domain).

| Prediction / Ground Truth | Upstairs | Downstairs | Negative | Accuracy |
|---|---|---|---|---|
| Upstairs | 79 | 1 | 0 | 98.75% |
| Downstairs | 0 | 79 | 1 | 98.75% |
| Negative | 0 | 0 | 63 | 100.0% |

Table.1 Experimental Results of the Escalator Staircase Recognition with the CNN architecture.

| Prediction / Ground Truth | Upstairs | Downstairs | Negative | Accuracy |
|---|---|---|---|---|
| Upstairs | 244 | 130 | 95 | 52.03% |
| Downstairs | 82 | 281 | 74 | 64.30% |
| Negative | 77 | 63 | 248 | 63.92% |

Table.2 Experimental Results of the Stationary Staircase Recognition with the CNN architecture **only** trained on the Escalator Dataset.

The results (as shown in Table.1) indicate that the proposed CNN can achieve an overall test accuracy of 99.28% on the source domain. After comparing the performance of the proposed CNN with the performance of Munoz et al's model (92.7%), we conclude that it is advantageous to use a deep CNN to extract discriminative features of the 4-channel RGB-D images. The results (overall 59.72%) shown in Table.2 indicate a significant degrade of performance when testing the same model on the target domain.

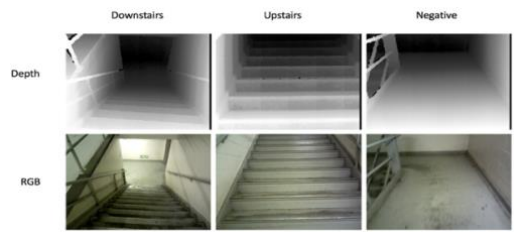

Fig.4 Different stationary staircases. The first row shows depth images; the second row shows the corresponding RGB images

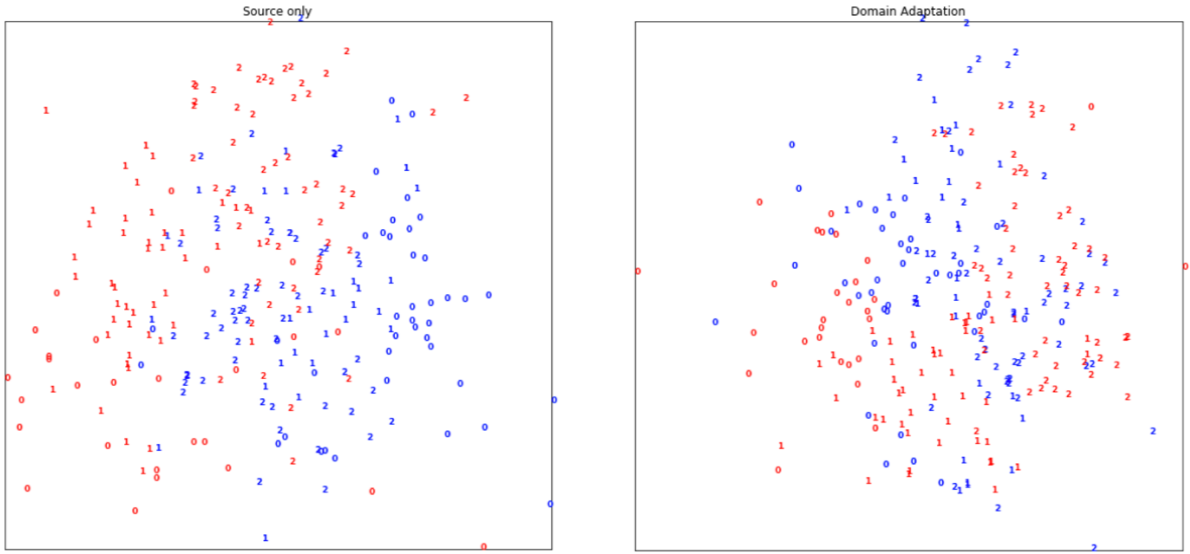

Fig.5 The effect of domain adaptation on the distribution of the extracted features. The left figure illustrates the feature distributions before domain adaptation. The right figure illustrates the adapted feature distributions. Blue points correspond to the escalator staircase data distributions (source domain). Red points correspond to the stationary staircase data distributions (target domain). The number 0 represents the downstairs case; the number 1 represents the upstairs case; the number 2 represents the negative case.

### 3.3 Results of Testing the Proposed Model with Domain Adaptation Regularizer

The proposed system incorporates the DANN regularizer with the proposed feed-forward CNN architecture. In order to mitigate the shift between two data distributions, we combined the labeled escalator staircase data with the **unlabeled** stationary staircase data to train the proposed system. Our model achieves an average accuracy at 100% for the escalator (source domain) staircase recognition. This result indicates that the domain adaptation regularizer can improve the performance of the classifier by adapting the escalator scenario to the stationary scenario.

| Prediction / Ground Truth | Upstairs | Downstairs | Negative | Accuracy |
|---|---|---|---|---|
| Upstairs | 80 | 0 | 0 | 100.0% |
| Downstairs | 0 | 80 | 0 | 100.0% |
| Negative | 0 | 0 | 63 | 100.0% |

Table.3 Experimental Results of the Escalator Staircase Recognition with Domain Adversarial Neural Networks.

After training convergence, we tested our model on the stationary staircase dataset (target domain). The results shown in Table 4 indicate that our model successfully mitigates the effect of the data distribution shift with an average test accuracy at 80.6% for the stationary staircase recognition.

| Prediction / Ground Truth | Upstairs | Downstairs | Negative | Accuracy |
|---|---|---|---|---|
| Upstairs | 381 | 56 | 32 | 81.24% |
| Downstairs | 47 | 339 | 51 | 77.57% |
| Negative | 25 | 40 | 323 | 83.25% |

Table.4 Experimental Results of the Stationary Staircase Recognition with Domain Adversarial Neural Networks.

### 3.4 Visualizations of Domain Adaptation

In order to visualize different feature distributions between the source domain and the target domain, we applied t-SNE projection [16] on the last hidden layer of the label predictor $G_y(f, \theta_y)$. As shown in Fig.5, the domain adaptation regularizer makes two different feature distributions get closer. This implies the feature extractor $G_f(., \theta_f)$ has been successfully confused by jointly training the label predictor $G_y(f, \theta_y)$ and the domain predictor $G_d(f, \theta_d)$.

The overlap between the different distributions in Fig.5 indicates the success of the adaptation. Moreover, we observed that the overlap also corresponds to the classification accuracy for the target domain, i.e., the more overlap in t-SNE projection, the higher classification accuracy for the target domain.

### 3.5 Computational Environment

The training and testing of the proposed model were implemented on a computer with an AMD Ryzen 7 2700X Eight-Core Processor (3.7 GHz), a 16 GB DDR3, and an NVIDIA GeForce GTX 1060 graphics card. This hardware environment allows the model to be trained on 256-sized batches. The labeled samples from the source domain take half of each batch. The rest of the batch constitutes the unlabeled samples from the target domain.

## 4 Discussion

In this research, we incorporate domain adaptation methods into staircase recognitions. In order to extract more transferable features from two different data distributions, we design a deep convolutional neural network to learn more discriminative features of the RGB-D staircase dataset. The results indicate that, without using domain adaptation techniques, our proposed CNN architecture can achieve 99.28% test accuracy on source domain data distributions (the escalator staircase data). Furthermore, our RGB-D staircase dataset has much less labeled data compared with Munoz et al.'s staircase dataset. However, our proposed CNN architecture still outperforms the model proposed by Munoz et al.

This research also indicates that domain adaptation methods will contribute to a better training convergence. By jointly training the label predictor with data from two different distributions, the model will outperform our proposed CNN architecture, and converge with fewer training epochs.

Although the proposed CNN architecture with domain adaptation regularizer can classify the staircase with high accuracy and mitigate the effect of data distribution shift, we acknowledge that there are still some limitations. Firstly, we should expand the categories of the staircase, such as obstacle, ramp, and wall, to enhance the environmental adaptability of the proposed model in more complex environments. Besides, we only adopt our model to two different staircase data distributions. More staircase data distributions should be considered in our future work so that our model will be more robust. Finally, the proposed method has only been evaluated in the offline analysis.

# 5 Conclusion

This paper presented a deep architecture for domain adversarial neural networks to transfer the knowledge learned from the labeled escalator staircase data distributions to the **unlabeled** stationary staircase data distributions. Unlike previous staircase classification methods, the accuracy of the proposed model does not rely on a large amount of labeled data. The results demonstrate that our model will achieve better performance with much less labeled data compared with other methods. Our model successfully mitigates the degrade of performance caused by the shift between data distributions. Moreover, our research indicates that RGB-D images have better feature representations than RGB images.

## Author Biographies

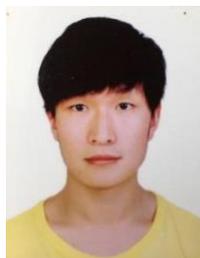

**JING Wang** received the B.ASc degree from The University of British Columbia (UBC) in 2018. Now he is an M.ASc student at the University of British Columbia. His research interests include computer vision and intelligent robotics.
Email: j.wang94@alumni.ubc.ca

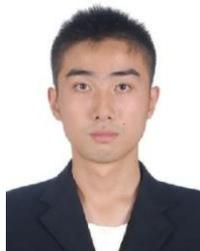

**KUANGEN Zhang** received the B.E. degree from Tsinghua University in 2016. Now he is a joint Ph.D. student of the University of British Columbia (UBC) and Southern University of Science and Technology (SUSTech). His research interests include robotic vision, sensor fusion, and wearable robots.
Email：kuangen.zhang@alumni.ubc.ca